\documentclass{article}
\usepackage{microtype}
\usepackage{graphicx}
\usepackage{subfigure}
\usepackage{booktabs} 

\usepackage{hyperref}

\usepackage[preprint]{icml2026}
\usepackage{amsmath}
\usepackage{amssymb}
\usepackage{mathtools}
\usepackage{amsthm}
\usepackage{enumitem}

\usepackage[capitalize,noabbrev]{cleveref}

\theoremstyle{plain}

\theoremstyle{definition}

\theoremstyle{remark}

\usepackage[textsize=tiny]{todonotes}

\icmltitlerunning{Subgoals as Compass: Temporal Vision-Language Reward Modeling for Complex Tasks}

\begin{document}

\twocolumn[
  \icmltitle{Subgoals as Compass: \\ Temporal Vision-Language Reward Modeling for Complex Tasks}



  \icmlsetsymbol{equal}{*}
    \begin{icmlauthorlist}
        \icmlauthor{Hyewon Choi}{equal,kaist}
        \icmlauthor{Minseok Jeong}{equal,kaist}
        \icmlauthor{SooJean Han}{kaist}
    \end{icmlauthorlist}
    \icmlaffiliation{kaist}{
        School of Electrical Engineering, KAIST,
        Daejeon, Republic of Korea
    }
    \icmlcorrespondingauthor{SooJean Han}{soojean@kaist.ac.kr}

  \icmlkeywords{Machine Learning, ICML}

  \vskip 0.3in
]



\printAffiliationsAndNotice{\icmlEqualContribution} 

\begin{abstract}
Reward specification remains a fundamental challenge, often requiring manual engineering or privileged state information. CLIP-based reward modeling offers a promising alternative by deriving rewards from visual observations and language descriptions, but existing methods often ignore temporal task structure or require additional training. In this paper, we propose a training-free vision-language model-based reward framework that enables sub-goal-aware reward construction under partial observability. Our method decomposes tasks into ordered sub-goals and estimates the current stage from visual observations using a CLIP-based adapter with a small support set. To improve reliability, we incorporate OOD detection to suppress unsupported reward assignments. We evaluate our approach on long-horizon MiniGrid tasks and show that it produces more structured reward signals and improves policy learning over existing reward formulations, suggesting a practical path toward scalable reward design.
\end{abstract}

\begin{figure*}
\centering 
\includegraphics[width=\textwidth]{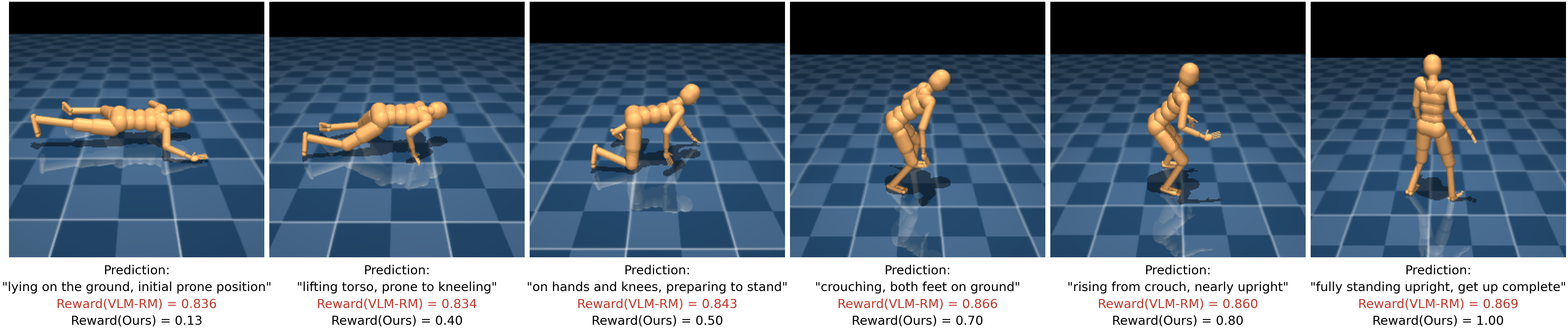}
\vspace{-1.2em}
\caption{\textbf{Why long-horizon rewards require temporal structure.}
A humanoid get-up motion requires progression through ordered intermediate sub-goals (lying, kneeling, crouching, and rising) before reaching the final goal of standing upright. Reward models that compute only a single similarity score with the final prompt fail to distinguish progress across these states.}
\label{fig:Reward_for_locomotion} 
\vspace{-0.1in} \end{figure*}
\section{Introduction}
Designing reward functions remains a fundamental bottleneck in reinforcement learning. While reward shaping based on low-level state variables is widely used (e.g., distance-to-goal metrics), such approaches require privileged access to ground-truth state unavailable in visual-only settings. A promising alternative is to derive reward signals directly from visual observations and natural language instructions using pretrained vision-language models (VLMs) such as CLIP~\cite{radford2021}. Recent work has explored this direction, with approaches measuring alignment between visual and textual embeddings or prompting VLMs to estimate task completion scores~\cite{rocamonde2023, wang2024}. However, VLM-based rewards are often highly sensitive to prompt design and poorly capture temporal progress in long-horizon tasks. Fig.~\ref{fig:Reward_for_locomotion} highlights this issue in a humanoid get-up task, where goal-only rewards fail to reflect ordered intermediate progress. Successful completion requires recognizing not only the final upright pose, but also transitions from \emph{lying prone} to \emph{fully standing upright}. This raises a critical question: How can we design a vision-based reward under partial observability that remains sub-goal aware? We argue that a reliable reward should satisfy two principles: \textbf{(i) Ordered Sub-goal Discrimination}—decompose the overall objective into an ordered sequence of sub-goals $(g_1,\dots,g_N)$ and assign rewards that reflect progression through them; \textbf{(ii) Avoiding Misleading Reward Signals}—mitigate perceptual aliasing, where visually similar observations correspond to different semantic sub-goals, by distinguishing in-distribution observations from out-of-distribution (OOD) ones.

Our contributions are summarized as follows:
First, we formulate long-horizon visual reward design as ordered sub-goal progression, converting a small set of labeled sub-goal observations into a dense reward signal.
Second, we demonstrate this idea using a training-free CLIP adapter, allowing few-shot sub-goal progression without reward-model fine-tuning.
Third, we evaluate the reward in MiniGrid DoorKey and show that it improves PPO learning over goal-only VLM rewards and state-based baselines; further experiments were performed on CartPole and humanoid environments.

\section{Related Work}
\subsection{Vision-Language Models for Reward  Generation}
Prior works have explored the use of VLMs to construct reward functions for reinforcement learning. A common approach is to measure alignment between observations and goal descriptions by computing the cosine similarity between visual representations and the embeddings of a text goal description~\cite{radford2021, rocamonde2023}. 
Some recent works directly prompt a VLM with an image observation and a textual goal to produce a scalar estimate of task completion~\cite{zhao2025}. 
Other approaches pose reward generation as a code synthesis problem, where LLMs generate executable reward functions from human feedback~\cite{venuto2024, kwon2023}. 
These approaches reduce manual reward engineering but remain prompt-sensitive and temporally inconsistent, often yielding flat rewards in long-horizon tasks.

\section{Preliminaries}
\subsection{Partially Observable Markov Decision Processes}
A partially observable Markov decision process (POMDP) is described by the tuple 
$(\mathcal{S}, \mathcal{A}, \mathcal{O}, \mathcal{P}, \Omega, r)$ 
where $\mathcal{S}$, $\mathcal{A}$, and $\mathcal{O}$ denote the state, action, and observation spaces, respectively. 
Here, $P(\cdot\mid s,a)\in\Delta(\mathcal{S})$ is the transition kernel, $\Omega(\cdot\mid s)\in\Delta(\mathcal{O})$ is the observation kernel, and $\Delta$ denotes the probability simplex over the given set.
The reward function $r : \mathcal{S} \times \mathcal{A} \times \mathcal{S} \to \mathbb{R}$ defines the reward received for each transition.

\subsection{Goal Decomposition in long-horizon tasks}
\label{sec:goal_decomposition}
Long-horizon tasks are naturally composed of multiple sub-goals \cite{okudo2021, yang2025, pertsch2020}. We therefore represent a task as an ordered sequence of sub-goals $\mathcal{G} = \{g_1, g_2, \ldots, g_N\}$, where each sub-goal corresponds to a meaningful stage of progress towards the final objective. Given an expert trajectory $\tau = \{(o_t, a_t, o_{t+1})\}_{t=1}^{T-1}$
we assign each transition to one of the $N$ sub-goal categories according to a predefined criterion $\mathcal C_n{\,=\,}\{o \in \mathcal{O} \mid o \text{ lies between sub-goals } g_{n-1} \text{ and } g_n\}$. The completion of sub-goal $g_n$ at time $t$ is represented by
\begin{equation}
\mathbb{I}_{\mathcal C_n}(o_t) =
\begin{cases}
1 & \text{if } {o_t} \in \mathcal C_n, \\
0 & \text{otherwise},
\end{cases}
\end{equation}
The observation-dependent reward is denoted by $\hat{r}$. The reward for a transition $(o_t, a_t, o_{t+1})$ can then be expressed as a weighted sum of sub-goal indicators:
$\hat{r}(o_t,a_t,o_{t+1})\triangleq\hat{r}(o_{t+1}) = \sum_{n=1}^{N} w_n \, \mathbb{I}_{\mathcal C_n}(o_{t+1})$. This decomposition reframes long-horizon reward design as sub-goal recognition, identifying the most likely sub-goal given the current observation.

\subsection{Tip-Adapter: Training-Free CLIP Adapter}
To recognize the sub-goal associated with a given observation using only a few datasets, we use a pretrained CLIP model as a frozen feature extractor and construct a training-free adapter on top of it. We build upon Tip-Adapter~\cite{zhang2021}, a training-free few-shot adaptation method for CLIP, and adopt its kernelized variant, ProKeR~\cite{bendou2025}, which performs adaptation in a Reproducing Kernel Hilbert Space (RKHS) with proximal regularization. This formulation enables adaptation from a small support set without additional training~\cite{zhou2022, gao2024}.

\begin{figure*}[t] 
\centering 
\includegraphics[width=0.95\textwidth, height=0.5\textwidth, trim=0 70 0 0, clip]{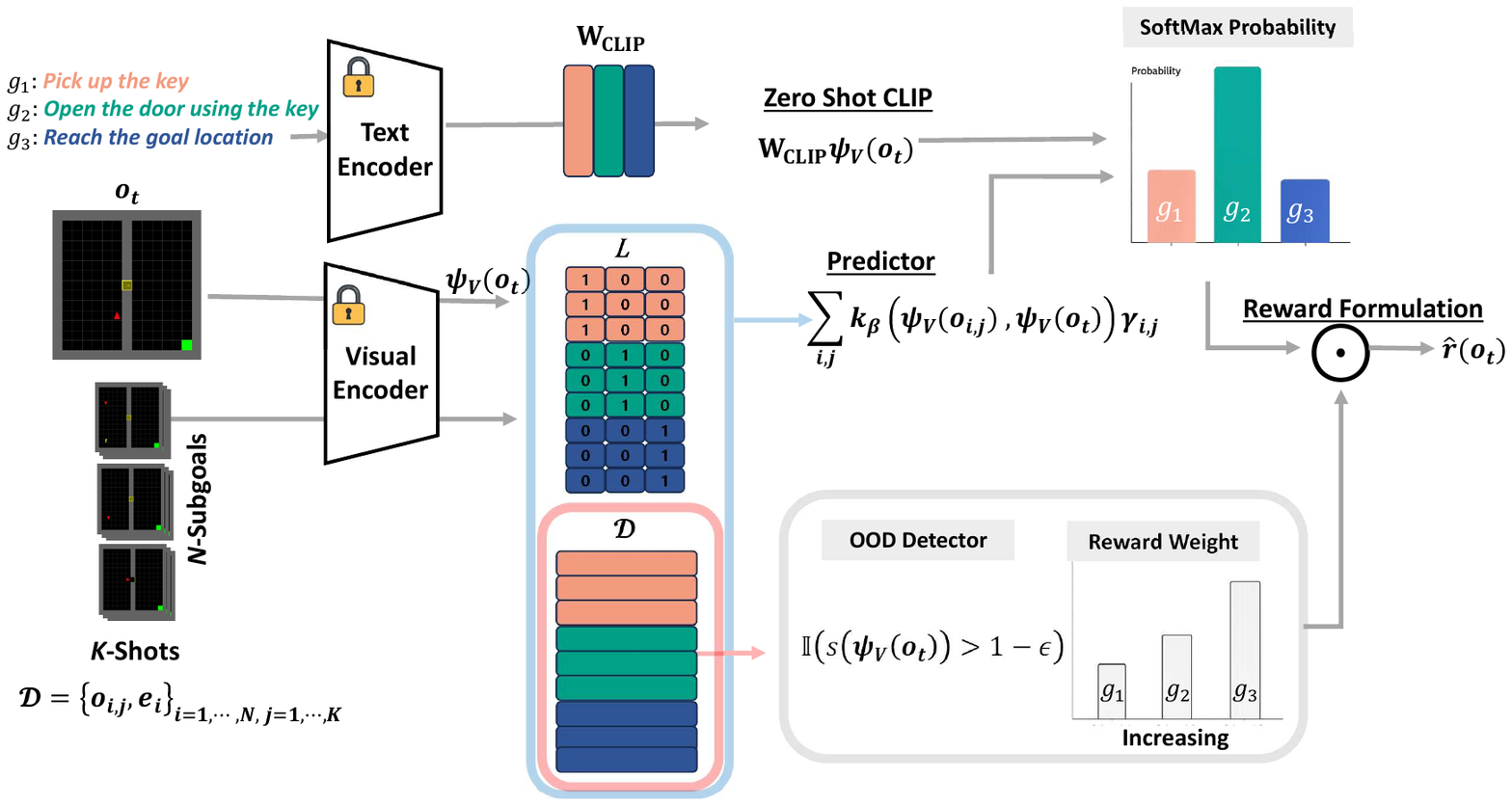}
\vspace{-1.0em}
\caption{Architecture for a subgoal-wise reward model using a pretrained vision-language model (VLM). The frozen VLM consists of an image encoder $\psi_V$ that extracts visual features $\psi_V(o_t)$ and a language encoder that produces text prototypes $W_{\text{CLIP}}$. Given the current observation $o_t$, these features are compared with both zero-shot text embeddings and a support set $\mathcal{D}$. A ProKeR-based predictor refines the CLIP logits via kernel regression over the support features. The resulting logits are converted into a probability distribution over sub-goals and weighted to produce a scalar reward. An OOD detector finally gates the output to yield the reward $\hat{r}(o_t)$.}
\label{fig:Overall Diagram} 
\vspace{-0.1in} \end{figure*}

Let $\psi_V(o) \in \mathbb{R}^D$ denote the visual feature of observation $o \in \mathcal{O}$ extracted by a frozen encoder and let $\psi_L(l_i) \in \mathbb{R}^D$ denote the text embedding of the $i$-th sub-goal description $l_i$. We define the text prototype matrix for $N$ classes as $W_{\mathrm{clip}}=[\psi_{L}(l_1),\dots,\psi_L(l_N)] \in \mathbb{R}^{D \times N}$, which yields the zero-shot predictor $f_{\text{clip}}(o) = \psi_V(o) W_{\mathrm{clip}}$. To adapt this predictor using a few labeled samples, we construct a support set
$\mathcal{D} = \{(o_{i,j}, e_i)\}_{i=1,\ldots,N;\; j=1,\ldots,K}$,
with $K$ samples per class, where $o_{i,j}$ is an observation from class $i$ and $e_i \in \mathbb{R}^N$ is its one-hot label. Given the support set, ProKeR defines the adapted predictor.\begin{align}
\phi(o) &= f_{\mathrm{clip}}(o) + \sum_{i=1}^{N}\sum_{j=1}^K k_\beta(\psi_V(o_{i,j}), \psi_V(o))\,\gamma_{i,j}, \\
\gamma &:= \left(I + \tfrac{1}{\lambda} K_\beta(\mathcal{D},\mathcal{D})\right)^{-1} \big(L - f_{\mathrm{clip}}(\mathcal{D})\big).
\end{align} Here, $k_\beta$ denotes a Gaussian kernel with bandwidth $\beta$, and $K_\beta(\mathcal{D}, \mathcal{D}) \in \mathbb{R}^{NK \times NK}$ is the corresponding Gram matrix computed over the support set. In addition, $f_{\mathrm{clip}}(\mathcal D) = [f_{\mathrm{clip}}(o_{1,1})^\top;\ldots;f_{\mathrm{clip}}(o_{N,K})^\top] \in\mathbb R^{NK\times N}$
and
$L=[e_1^\top;\ldots;e_1^\top;\ldots;e_N^\top;\ldots;e_N^\top]
\in\mathbb R^{NK\times N}$
for the stacked zero-shot logits and one-hot labels, respectively.

\section{Method}
We assume access to expert demonstration trajectories sampled from $p_E$, where each trajectory is given by
$\tau = \{(o_t, a_t, o_{t+1})\}_{t=1}^{T-1}$. Our goal is to construct a vision-based reward function that satisfies the two key requirements outlined above, without additional training. The overall framework is illustrated in Fig.~\ref{fig:Overall Diagram}.

\subsection{Support Set Construction}
To build the support set, we partition expert trajectories into $N$ sub-goal categories following the criterion defined in Sec.~\ref{sec:goal_decomposition}. For each sub-goal $i \in \{1, \ldots, N\}$, we select $K$ representative frames $\{o_{i,1}, o_{i,2}, \ldots, o_{i,K}\}$ that capture the temporal progression and transitions within that phase. Each observation $o_{i,j}$ is then paired with a one-hot label $e_i \in \mathbb{R}^N$ that indicates which sub-goal it belongs to. Collecting these labeled examples across all stages yields the support set. 
\subsection{Reward Specification}
\label{sec:grounding}
Given an observation $o$, we use ProKeR to estimate which sub-goal stage it most likely corresponds to. The CLIP adapter 
produces logits $\phi(o) \in \mathbb{R}^N$, where each entry corresponds to one of the $N$ ordered sub-goals. We convert these logits into a probability distribution over sub-goal stages using temperature-scaled softmax: $p(n \mid o) = \mathrm{softmax}\big(T \cdot \phi(o)\big)_n$, where $n \in \{1,\ldots,N\}$ and $T>0$ controls the sharpness of the distribution. We then define the reward as a weighted expectation over task stages:
\begin{equation}\label{eq:reward_weighted}
    \hat{r}(o) = \sum_{n=1}^{N} w_n\, p(n \mid o),
\end{equation}
where the stage weights satisfy $w_1 \leq w_2 \leq \cdots \leq w_N$. Because later sub-goals receive larger weights, the reward increases as the observation is assigned to later stages.

\subsection{Out of Distribution (OOD) Detector}
To prevent assigning incorrect rewards to unsupported observations, we apply OOD detection using a pretrained encoder. We compute the cosine similarity between the L2-normalized feature and the support set as $s(\psi_{V}(o))=\max_{i\in\{1,\dots,KN\}} \psi_{V}(o)^\top \mathbf{v}_i.$
Based on this similarity, we perform an ID/OOD decision using a threshold $\epsilon$, i.e.,
$\mathbb{I}\left[s(\psi_{V}(o)) \geq 1 - \epsilon\right]$.
OOD observations are assigned zero reward and treated as unexplored. In practice, this component can be instantiated using standard approaches such as one-class SVM, kNN-based detectors, or Random Network Distillation (RND)~\cite{Ramaswamy2000, scholkopf2001, burda2018}.

\section{Experiments}
\begin{figure}[t]
    \centering
    \subfigure[]{
        \includegraphics[width=0.44\linewidth]{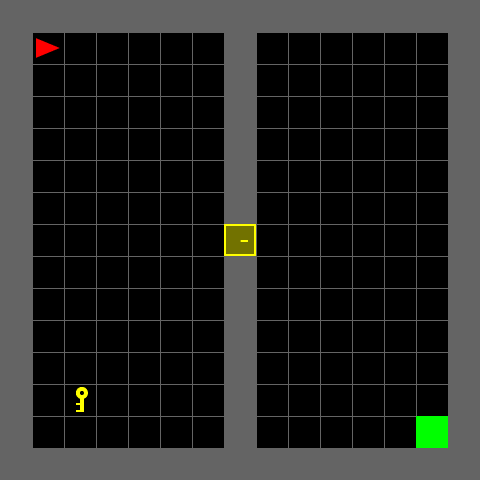}
    }
    \hfill
    \subfigure[]{
        \includegraphics[width=0.46\linewidth]{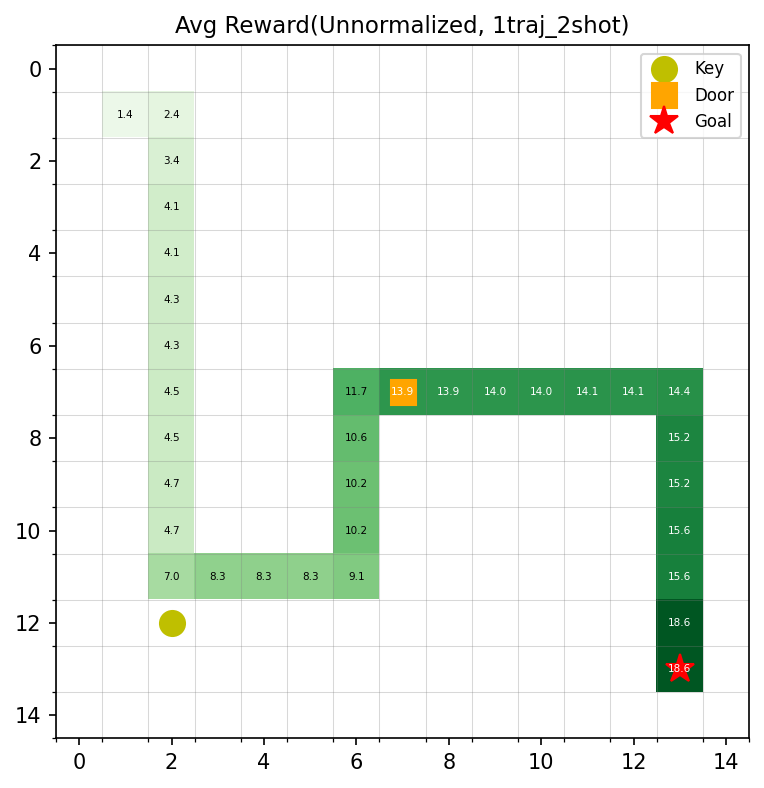}
    }
    \caption{Left: MiniGrid DoorKey task. Right: Reward heatmap computed from cached support features along the expert trajectory. The reward increases as the agent progresses through subgoals (key, door, goal).}
    \label{fig:reward_heatmap}
\end{figure}

\paragraph{Environment.} We evaluate our method on the DoorKey task (Fig.~\ref{fig:reward_heatmap}(a)) in a 2D MiniGrid environment.
We distinguish high-level semantic sub-goals from ordered sub-goals; here, the semantic sub-goals are
picking up a key, opening a door, and reaching the goal. During training, the agent receives RGB observations of size $84 \times 84 \times 3$. 
\paragraph{Reward Construction.} For each expert trajectory, we discretize $N=20$ ordered sub-goals and assign monotone weights as prescribed by~\eqref{eq:reward_weighted}.
Each sub-goal is represented by $K{=}2$ support observations (shots) and we assign a text prototype to each sub-goal class. We use ViT-bigG-14 as the CLIP backbone. Unsupported features are treated as OOD and assigned zero reward. During training, the assigned rewards are normalized by the maximum reward value. Fig.~\ref{fig:reward_heatmap}(b) illustrates the temporal reward across sub-goals assigned by our model. 
\paragraph{Training Setup}
The policy is trained using Proximal Policy Optimization (PPO) with RGB observations. Training details are provided in Table~\ref{tab:train_details}. For comparison, we consider four reward baselines:
\begin{itemize}[topsep=0pt]
    \item \textbf{Event-based State Reward}: assigns rewards only when predefined task events are completed (key pickup, door opening, goal reaching),  $\hat r(o)=\sum_{n=1}^{N} w_n\,\mathbb{I}_{\mathcal C_n}(o)$, $o \sim \Omega(\cdot \mid s).$ 
    \item \textbf{Distance-based State Reward}: shapes rewards according to the Mahalanobis distance to each sub-goal $d_n(s)$, defined as $r(s)=\sum_{n=1}^{N} w_n\, [\exp( -d_n(s)) \mathbb{I}_{\mathcal{C}_n}(o)]$, $o \sim \Omega(\cdot \mid s)$.
    \item \textbf{Goal-alignment Reward (VLM-RM)}: measures alignment with the final goal description $l_N$ by projecting visual features onto the goal direction, $\hat r_{N, \mathrm{VLM-RM}}(o)=1-\frac{1}{2}\|\operatorname{proj}_{\psi_L(l_N)}(\psi_V(o))-\psi_L(l_N)\|_2^2$, following \cite{rocamonde2023}.
    \item \textbf{Temporal Goal-alignment Reward (VLM-RM (temporal))}: extends VLM-RM by aggregating scores over decomposed sub-goal prompts, $\hat r(o)=\sum_{n=1}^{N} w_n\, \hat r_{n, \mathrm{VLM-RM}}(o)$. 
\end{itemize}
\begin{figure}[t]
\centering 
\includegraphics[width=\columnwidth]{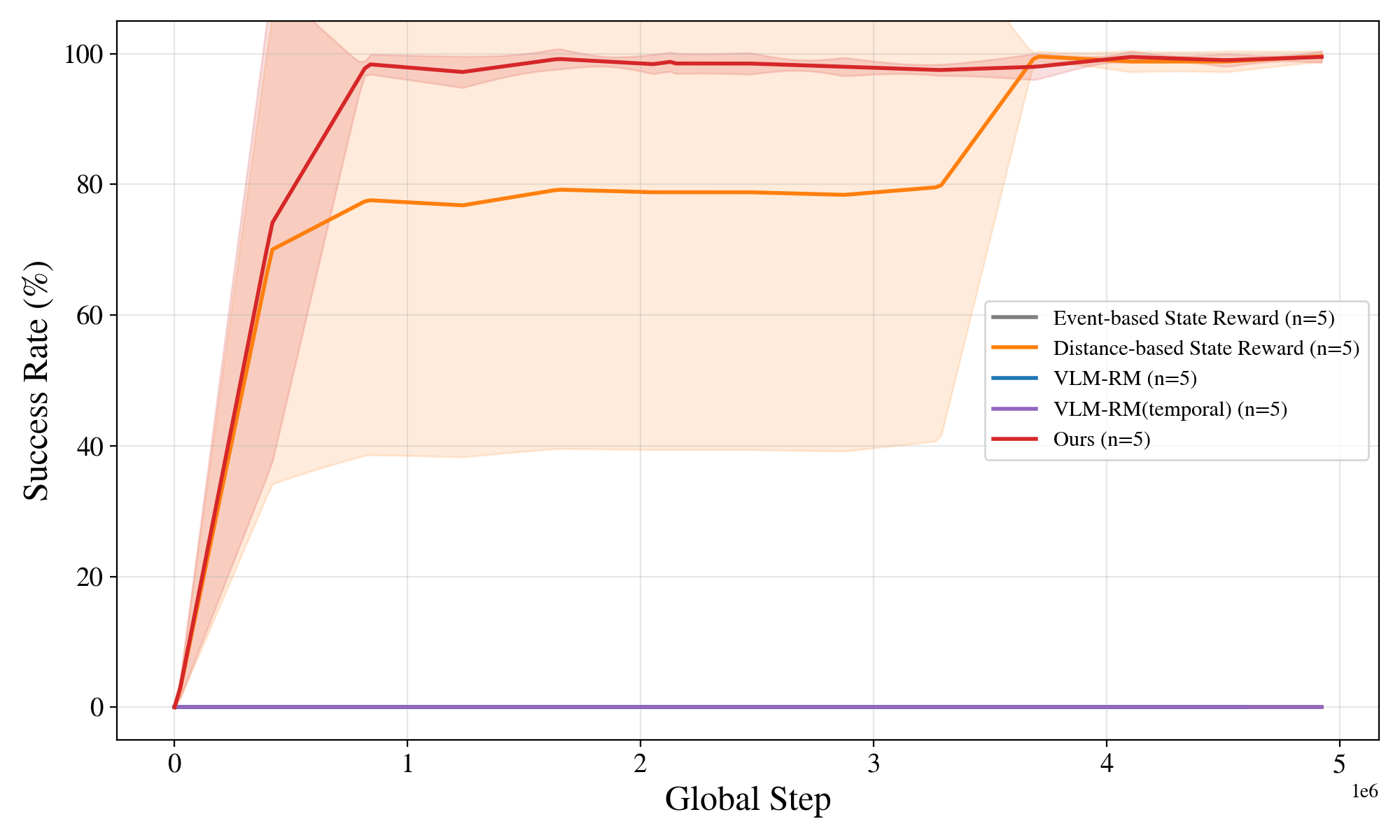}
\vspace{-1.2em}
\caption{Success rates of PPO on MiniGrid under five different reward model variants. Our stage-wise reward formulation achieves the highest final performance and more stable learning compared to alternatives}
\label{fig:success_rate} 
\vspace{-0.1in}
\end{figure}
\subsection{Main Results: Online RL Performance Across Reward Designs}
We present online RL results in Fig.~\ref{fig:success_rate}. Across all baselines, our reward achieves the fastest and most stable learning, reaching near-perfect success within $10^6$ global steps. In contrast, the distance-based state reward improves more slowly and requires more interaction to achieve comparable final performance. Interestingly, the goal-alignment reward (VLM-RM) does not acquire a meaningful exploration from the goal alignment alone, while the failure of VLM-RM (temporal) suggests that effective reward design requires more than text-image alignment. Capturing fine-grained visual differences across subgoals is also critical for successful policy learning.

\section{Limitation and Future Works}
A limitation appears in the humanoid setting, where high-dimensional observations and continuous control hinder learning. Although the reward model produces plausible subgoal signals (Fig.~\ref{fig:Reward_for_locomotion}), RGB-based PPO fails to learn a successful policy in the humanoid setting. In particular, OOD detection fails to filter perceptually similar but irrelevant states, leading to misleading rewards, while exploration is further challenged by the large state space. Future work will focus on more robust OOD detection and improved exploration in complex continuous-control domains.

\section{Conclusion}
We presented a training-free vision-language reward framework for RL under partial observability. Our method reformulates reward design as ordered sub-goal recognition using a CLIP-based few-shot adapter, with OOD detection to improve reliability. Experiments on MiniGrid show more informative reward signals and improved policy performance over goal-only and state-based baselines. Additional results suggest applicability to continuous-control domains and that multi-trajectory support sets improve reachability. Overall, our findings highlight the potential of pretrained vision-language models as scalable reward models without dense state supervision.

\section*{Impact Statement}
This work studies how pretrained vision-language models can turn weak and partially observable environmental signals into practical reward feedback for complex decision-making. Instead of relying on handcrafted state-based rewards or costly human preference annotation, our approach uses a small set of labeled subgoal image-text pairs with lightweight adaptation to infer task progress from observations. In domains where reward specification is difficult, such as robotics or visually grounded control, this direction may reduce reward engineering costs while providing more scalable feedback signals.

\bibliography{example_paper}
\bibliographystyle{icml2026}


\newpage
\appendix
\onecolumn
\section{Appendix}
\begin{table*}[t]
\caption{Training Configuration Details for CLIP and PPO in MiniGrid.}
\label{tab:train_details}
\vskip 0.15in
\begin{center}
\begin{small}
\begin{sc}
\begin{tabular}{p{4.0cm}p{7.5cm}}
\toprule
\multicolumn{2}{c}{\textbf{CLIP Configuration}} \\
\midrule
\textbf{CLIP backbone} & ViT-bigG-14 (laion2b\_s39b\_b160k) \\
\textbf{Few-shot (k)} & 2 \\
\textbf{Best $\beta$ (RBF kernel)} & 3.0 \\
\textbf{Best $\lambda$ (regularization)} & 0.527 \\
\textbf{Temperature (T)} & 50 \\
\textbf{Reward weights} & $\{0, 1, 2, 3, 4, 5, 6, 7, 8, 9, 10, 11, 12, 13, 14, 15, 16, 17, 18, 19\}$ \\
\textbf{Reward Normalization} & True \\
\bottomrule
\end{tabular}
\vskip 0.15in
\begin{tabular}{p{4.0cm}p{7.5cm}}
\toprule
\multicolumn{2}{c}{\textbf{PPO Training Configuration}} \\
\midrule
\textbf{Algorithm} & RGB-Based PPO \\
\textbf{Policy network} & NatureCNN (RGB) + MLP \\
\textbf{CNN architecture} & 3×Conv2d (32→64→64) + FC(256) + ReLU \\
\textbf{Actor/Critic hidden} & 512 \\
\textbf{Total timesteps} & 10,000,000 \\
\textbf{Num envs (training)} & 128 \\
\textbf{Num envs (eval)} & 4 \\
\textbf{Rollout length} & 50 steps \\
\textbf{Batch size} & 128 × 50 = 6,400 \\
\textbf{Minibatches} & 8 \\
\textbf{Update epochs} & 8 \\
\textbf{Learning rate} & $3 \times 10^{-4}$ \\
\textbf{Discount ($\gamma$)} & 0.8 \\
\textbf{GAE $\lambda$} & 0.9 \\
\textbf{Target KL} & 0.2 \\
\textbf{Value loss coefficient} & 0.5 \\
\textbf{Entropy coefficient} & 0.0 \\
\textbf{Max grad norm} & 0.5 \\
\textbf{Eval frequency} & Every 10 iterations \\
\bottomrule
\end{tabular}
\end{sc}
\end{small}
\end{center}
\vskip -0.1in
\end{table*}
\subsection{Generalization to Diverse Control Environments}
\begin{figure}[H]
    \centering  
    \includegraphics[width=0.5\textwidth]{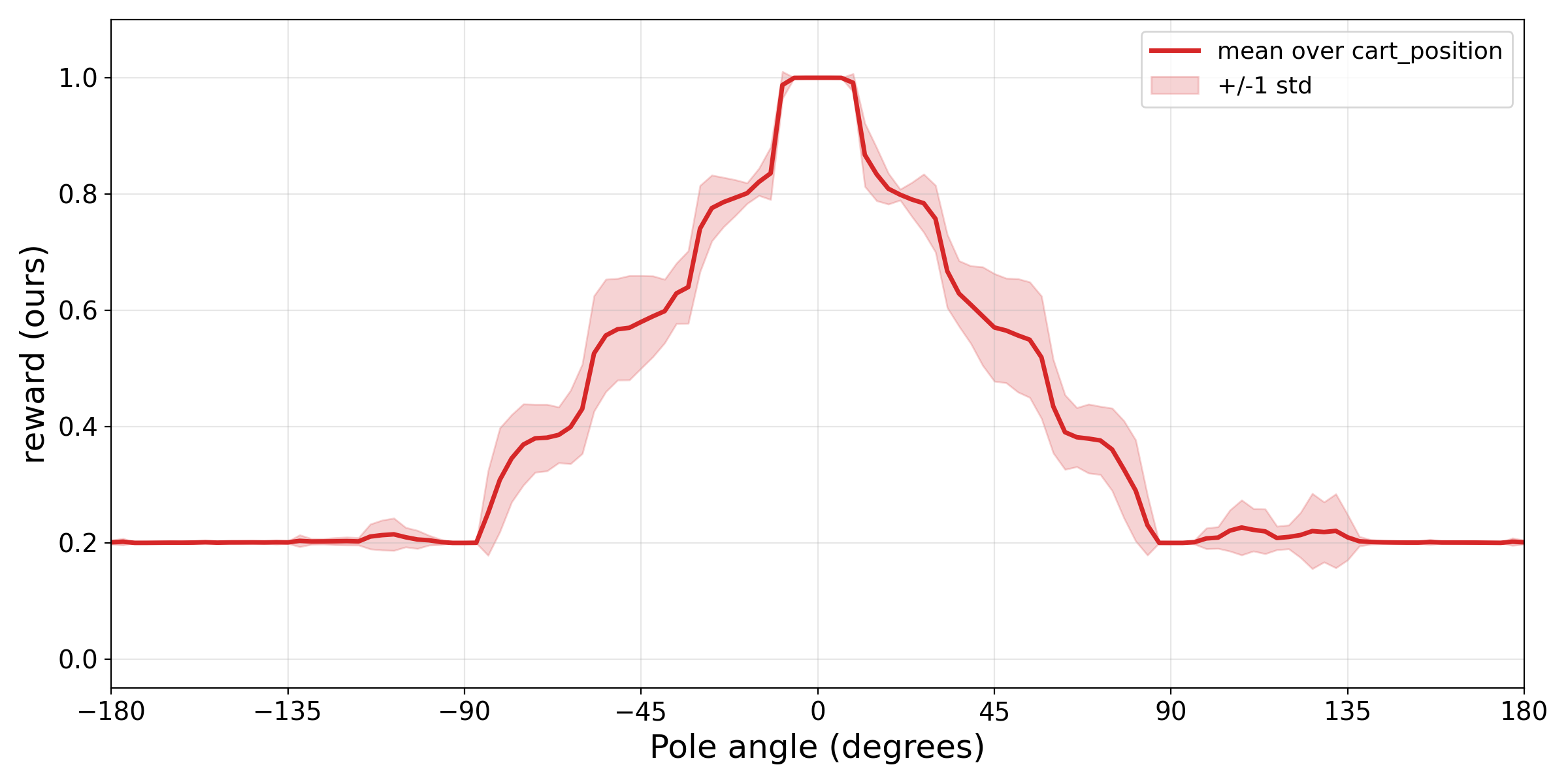}
    \caption{
    Reward profile across pole angles in the CartPole environment using five semantic classes: \textit{diagonal}, \textit{lying flat}, \textit{moderately to the side}, \textit{nearly flat}, and \textit{straight upward}. Higher rewards are assigned as the pole approaches the upright configuration. The curve denotes the mean reward and the shaded region indicates the standard deviation over different cart positions.
    }
    \vspace{-.4cm}
    \label{fig:cartpole_reward_eval}
\end{figure}
To evaluate the scalability and robustness of the proposed reward model, we extend our experiments to additional control benchmarks: \texttt{Cartpole} and \texttt{Humanoid} from the MuJoCo/DM Control suite. In the CartPole environment, we evaluate whether the proposed reward captures the semantic progression of pole configurations. Pole angles from $0^\circ$ to $90^\circ$ are divided into five semantic classes: \textit{diagonal}, \textit{lying flat}, \textit{moderately to the side}, \textit{nearly flat}, and \textit{straight upward} --- with a 10-shot support set per class. As shown in Fig.~\ref{fig:cartpole_reward_eval}, the learned reward increases consistently as the pole approaches the upright configuration. The solid curve denotes the mean reward, while the shaded region indicates the standard deviation over different cart positions. For the humanoid get-up task, we construct a support set from a single humanoid from \texttt{dm\_control} with a single motion capture trajectory, partitioned into $N{=}10$ ordered sub-goals ranging from \emph{lying prone} to \emph{standing upright}. For each subgoals, we sample $K{=}10$ representative observations, yielding a total of 100 support images. We evaluate generalization on unseen observations that are not included in the support set. To isolate the sub-goal recognition capability of the CLIP adapter, OOD gating is disabled in this experiment. As shown in Fig.~\ref{fig:Reward_for_locomotion}, the model correctly predicts the sub-goal and produces a monotonically increasing reward across the trajectory. These results suggest that the proposed method can capture meaningful task progression even in high-dimensional continuous-control settings.

\subsection{Does multi-trajectory reward construction improve reachability?}
The construction of Single-trajectory reward often induces path-specific bias, limiting generalization to unseen states. To mitigate this issue, we construct the support set from multiple expert trajectories, providing broader coverage of the state space. We then perform a recovery experiment that measures whether the resulting value function can guide the agent to the goal from randomly initialized states. To probe the geometry of the learned value landscape, we construct the action-value function using a pretrained critic $V_\theta$ via a one-step Bellman backup:
\begin{equation}
Q(o,a)=\hat{r}(o,a,o')+\rho V_\theta(o') = \hat{r}(o')+\rho V_\theta(o'),
\end{equation}
where $o \in \mathcal{O}$ denotes the current observation, $a \in \mathcal{A}$ the action, $o'$ the next observation, $\hat{r}(o')$ the immediate reward, and $\rho \in [0,1)$ the discount factor. We then sample actions from a Boltzmann recovery policy:
\begin{equation}
\pi_{\mathrm{recovery}}(a \mid o)
=
\frac{\exp(Q(o,a))}
{\sum_{a'} \exp(Q(o,a'))}.
\end{equation}

\begin{figure}[t]
    \centering    
    \begin{minipage}[t]{0.3\textwidth}
        \centering
        \includegraphics[width=\linewidth]{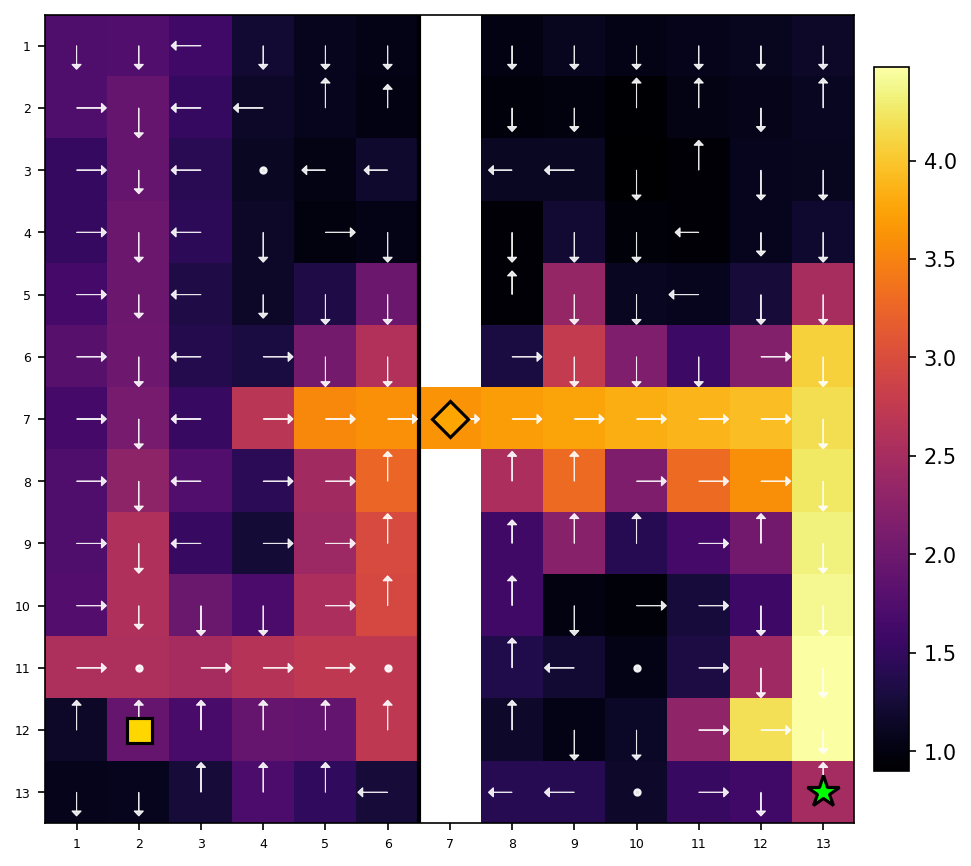}
        {\tiny (a) 1 trajectory, 2-shot}
    \end{minipage}
    \hspace{0.015\textwidth}
    \begin{minipage}[t]{0.3\textwidth}
        \centering
        \includegraphics[width=\linewidth]{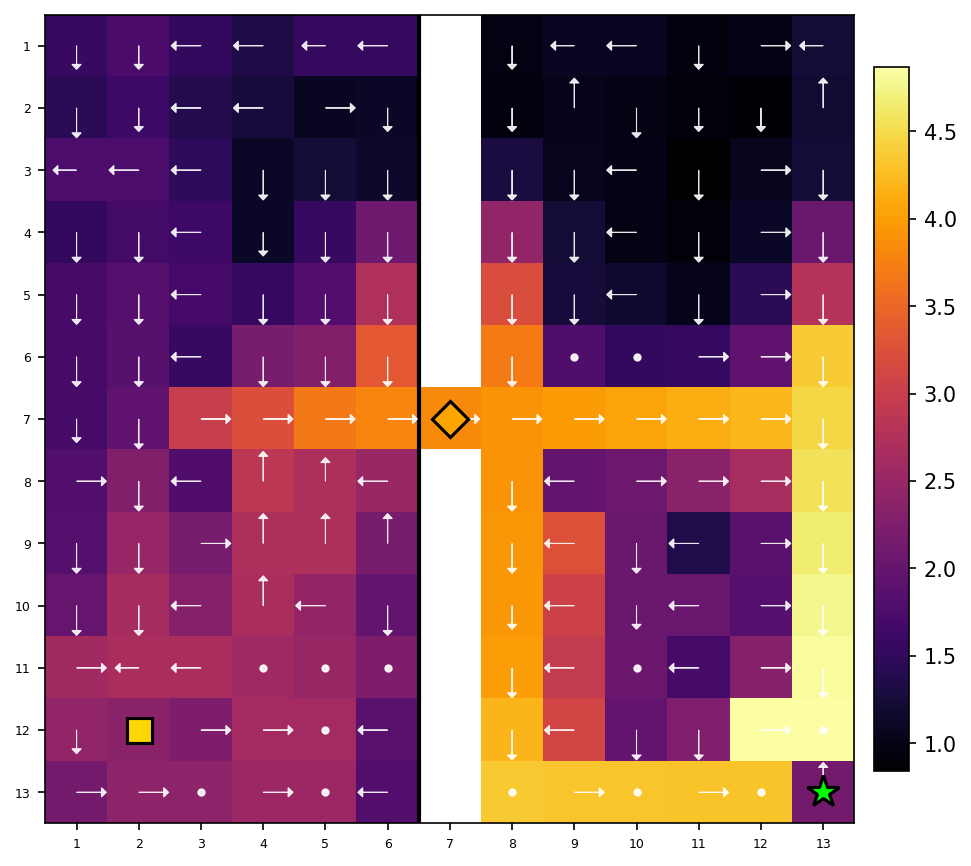}
        {\tiny (b) 3 trajectories, 6-shot}
    \end{minipage}
    \caption{Approximated $Q$-value landscapes learned from reward distributions. Unlike the single-trajectory variant (a), which contains dark dead zones where value gradients provide no guidance, the multi-trajectory setting (b) establishes smooth value gradients that enable reachability even from states far from the expert trajectories, allowing the agent to navigate toward the goal from arbitrary initial positions.}
    \vspace{-.3cm}
    \label{fig:qvalue_heatmap}
\end{figure}

Our core hypothesis is that multi-trajectory reward construction induces a value landscape that captures the geometric structure of \textit{reachability}, enabling robust goal-reaching behavior from diverse initial states. As shown in Fig~\ref{fig:qvalue_heatmap} visualizes the learned $Q$-value landscapes for single-trajectory (1 trajectory, 2-shot) versus multi-trajectory (3 trajectories, 6-shot) settings on MiniGrid. The 3-trajectories exhibits a substantially smoother and more coherent value landscape guiding toward the goal, whereas the single-trajectory construction shows fragmented high-value regions with dead zones (dark color). As shown in Table~\ref{tab:recovery_q}, this improved landscape structure translates to better recovery performance: the agent successfully reaches the goal from diverse unseen initial states within the maximum under the 3-trajectory setting, suggesting that multi-trajectory learning captures generalizable reachability.

\begin{table}[H]
\caption{Recovery Analysis. Entries report the recovery rate of reaching the final goal within the specified Max Steps, from initial states sampled uniformly across three conditions (333 each) ($N{=}999$, $\rho{=}0.8$).}
\label{tab:recovery_q}
\vskip 0.15in
\begin{center}
\begin{small}
\begin{sc}
\setlength{\tabcolsep}{20pt}
\begin{tabular}{lcc}
\toprule
Max Steps & 1 Traj & 3 Trajs \\
\midrule
100  & 17.8\%   & \textbf{22.9}\%  \\
500  & 64.2\% & \textbf{71.2}\%  \\
1000 & 83.6\% & \textbf{90.7}\%  \\
\bottomrule
\end{tabular}
\end{sc}
\end{small}
\end{center}
\end{table}
\end{document}